\pdfoutput=1

\documentclass[11pt]{article}

\usepackage[final]{acl}

\usepackage{times}
\usepackage{latexsym}

\usepackage[T1]{fontenc}

\usepackage[utf8]{inputenc}

\usepackage{microtype}

\usepackage{inconsolata}

\usepackage{graphicx}

\usepackage{array}
\newcolumntype{L}[1]{>{\raggedright\arraybackslash}p{#1}}
\usepackage{booktabs}
\usepackage{amsmath}

\newcommand{\Sref}[1]{\S\ref{#1}}

\title{How do we measure privacy in text? A survey of text anonymization metrics}

\author{Yaxuan Ren \and Krithika Ramesh \and Yaxing Yao \and Anjalie Field \\
Johns Hopkins University \\
\texttt{\{yren46, kramesh3, yaxing, anjalief\}@jhu.edu}
}

\begin{document}
\maketitle
\begin{abstract}
In this work, we aim to clarify and reconcile metrics for evaluating privacy protection in text through a systematic survey.
Although text anonymization is essential for enabling NLP research and model development in domains with sensitive data, evaluating whether anonymization methods sufficiently protect privacy remains an open challenge. In manually reviewing 47 papers that report privacy metrics, we identify and compare six distinct privacy notions, and analyze how the associated metrics capture different aspects of privacy risk. We then assess how well these notions align with legal privacy standards (HIPAA and GDPR), as well as user-centered expectations grounded in HCI studies. Our analysis offers practical guidance on navigating the landscape of privacy evaluation approaches further and highlights gaps in current practices. Ultimately, we aim to facilitate more robust, comparable, and legally aware privacy evaluations in text anonymization.

\end{abstract}

\section{Introduction}

Text anonymization—through methods such as redaction, rewriting, or data synthesis—has become a critical tool for mitigating the risks of sharing or training models on sensitive data \citep{lison_anonymisation_2021}. When done effectively, anonymization can enable access to valuable resources like clinical records, legal texts, or social media content without compromising individual privacy. However, text anonymization is inherently difficult: high-dimensional data in general is vulnerable to re-identification \citep{narayanan_robust_2008}, and even mechanisms that offer formal guarantees can still fail against practical deanonymization attacks \citep{mattern_limits_2022, tong_vulnerability_2025, pang2025reconstructiondifferentiallyprivatetext}.
As LLMs further heighten concerns about memorization of training data \citep{carlini_extracting_2021} and re-identification of sensitive attributes \citep{staab_beyond_2024}, rigorous privacy evaluation has become a fundamental requirement for responsible data sharing and model deployment.

Despite the importance of evaluation, measuring the effectiveness of text anonymization systems remains an open challenge. Current evaluations span a wide range of tasks and assumptions, reflecting divergent notions of privacy. Papers focusing on redacting direct identifiers \citep{hassan_automatic_2019, lison_anonymisation_2021, pilan_text_2022} implicitly use different notions of privacy than papers focusing on synthesizing text \citep{meisenbacher_1-diffractor_2024, yue_synthetic_2023, wang_differentially_2023}, and even papers targeting the same notion of privacy use different metrics.
In many cases, metrics are poorly connected to legal and social notions of privacy, leaving researchers and practitioners with limited guidance on what risks a given evaluation actually measures, or what constitutes sufficient protection in practice.

In this work, we aim to enhance the understanding of privacy evaluation in text by conducting a systematic survey of metrics.
Through keyword searching and citation links, we identify 47 papers published since 2019 that report metrics for measuring privacy in text, and we manually categorize metrics into high-level notions of privacy.
Unlike existing surveys of anonymization techniques \citep{pawar_anonymization_2018, mahendran_review_2021} or general privacy principles \citep{wagner_technical_2019}, we specifically target quantified metrics.
While we focus primarily on metrics for evaluating privacy in text that has been anonymized (e.g., through redaction, rewriting, or synthesis), our analysis also has relevance to privacy in models trained on text.


Our analysis reveals six privacy notions underlying specific metrics—identifier removal effectiveness, dataset membership, attribute inference risk, reconstruction attacks, semantic inference risk, and theoretical privacy bounds, which we discuss in more depth in \Sref{sec:survey}. We further discuss how these notions map to legal privacy standards, specifically HIPAA and GDPR (\Sref{sec:legal}), and social expectations derived from user-centered research (\Sref{sec:user}). We further relate these findings to model-privacy research and argue for bridging text and model privacy in \Sref{sec:discussion}. Finally, we conclude by discussing open challenges in privacy evaluation and opportunities for future work (\Sref{sec:rec}).

Overall, we aim to help researchers and practitioners understand the landscape of privacy evaluations for text, offering a structured view of what existing metrics capture and what they overlook. By organizing metrics by privacy objectives and examining their assumptions, we offer practical guidance for metric selection and reveal gaps in legal alignment, social relevance, and evaluation consistency. We ultimately aim to improve the consistency in evaluation and encourage future work on the development of new standardized metrics.

\section{Existing privacy metrics in NLP}
\label{sec:survey}
\subsection{Scope and Methodology}
We define the scope of this survey as metrics for evaluating privacy in anonymized text outputs, i.e., evaluations that directly assess how much sensitive information remains exposed in the text itself after anonymization. Our goal is to characterize the landscape of privacy evaluations applied to generated or modified text, focusing on settings where both original and anonymized versions are typically available.

We include papers published from 2019 onward that explicitly report one or more quantitative metrics for evaluating privacy in anonymized or synthetic text. This time window ensures our review focuses on recent methods relevant to current NLP pipelines and data-sharing concerns. We identified papers using a combination of keyword-based search and backward citation tracking. Specifically, we searched ACL Anthology and Google Scholar using combinations of the terms \texttt{``text anonymization''}, \texttt{``text sanitization''}, and \texttt{``synthetic text generation''}. These searches returned thousands of results (for example, over 1,500 and 16,000 papers respectively for ``text anonymization'' on ACL Anthology and Google Scholar), though most were only tangentially related to privacy evaluation. We therefore manually screened results for relevance and retained 47 papers that satisfied our inclusion criteria.

To be included, a paper must satisfy all of the following:
\begin{itemize}
  \item Focus on natural language text (not images, structured tabular data, or speech);
  \item Contain anonymized or privatized text outputs (not just internal model embeddings or representations);
  \item Report at least one privacy evaluation metric (beyond utility, fluency, or readability);
  \item Use primarily English-language data;\footnote{We excluded papers focusing exclusively on non-English datasets due to differing annotation schemes and accessibility.}
  \item Be peer-reviewed or publicly available as a preprint since January 2019.
\end{itemize}

\subsection{Survey of Evaluation Metrics}
\label{sec:survey}

\begin{figure*}[h]
    \centering
    \includegraphics[width=\textwidth]{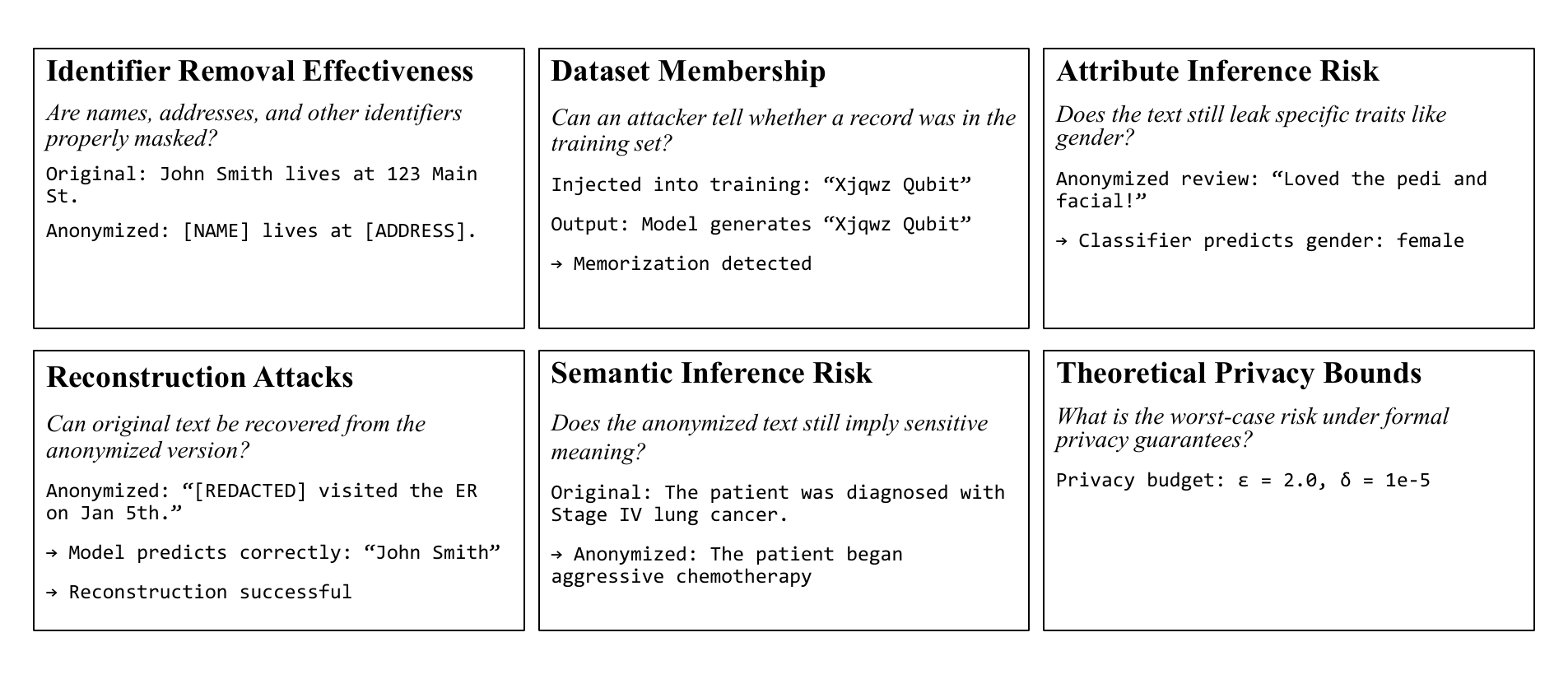}
    \caption{Overview of six privacy objectives used in text privacy evaluation. Each panel summarizes the privacy notion and provides an illustrative example.}
    \label{fig:privacy_objective}
\end{figure*}

We identify six high-level privacy objectives that the surveyed papers aim to evaluate: identifier removal, dataset membership, attribute inference, reconstruction attacks, semantic inference, and theoretical bounds. Figure~\ref{fig:privacy_objective} provides an overview of these objectives, and Table~\ref{tab:rep_metric_by_objective} provides representative examples of metrics and methods used to assess each objective. Each objective reflects a different aspect of privacy risk and corresponds to distinct families of evaluation metrics. Some papers target a single objective, while others report metrics spanning multiple categories. In the subsections that follow, we describe each objective in turn, outline the types of metrics used to evaluate it, and discuss how these metrics are applied in practice. A detailed mapping of papers to privacy objectives and associated evaluation metrics is provided as a spreadsheet in our GitHub repository.\footnote{\url{https://github.com/ryxGuo/privacy-metrics-survey}}.

\begin{table*}[h]
\centering
\small
\setlength{\tabcolsep}{4pt}
\renewcommand{\arraystretch}{1.2}
\begin{tabular}{L{1.8cm} L{0.4cm} L{4.5cm} L{2cm} L{2.2cm} L{2cm}}
\toprule
\textbf{Objective} & \textbf{\#} & \textbf{Representative Metric} & \textbf{Paper} & \textbf{Anonymization Method} & \textbf{Dataset} \\
\midrule
Identifier Removal Effectiveness & 18 & Precision, Recall, F1-score: standard detection metrics for correct vs.\ missed redactions. \(F_1 = 2PR/(P+R)\)
 & \citet{lison_anonymisation_2021} & Presidio & Wikipedia biographies \\
 \midrule
Dataset Membership & 8 & Membership Inference Attack (MIA) AUC: area under ROC curve measuring how well an attacker distinguishes members vs.\ non-members. & \citet{arnold_guiding_2023} & Synthetic text generation & IMDb \\
\midrule
Attribute Inference Risk & 9 & Attribute inference attack success rate: fraction of anonymized samples where sensitive attributes remain correctly predicted. & \citet{wang_promptehr_2022} & PromptEHR & MIMIC-III \\
\midrule
Reconstruction Attacks & 16 & Text Re-Identification Risk (TRIR): proportion of anonymized texts whose originals are correctly retrieved. & \citet{pilan_truthful_2024} & INTACT & Text Anonymization Benchmark (TAB) \\
\midrule
Semantic Inference Risk & 8 & BLEU score: $n$-gram overlap between original and anonymized text (higher = more semantic similarity). & \citet{igamberdiev_dp-bart_2023} & ADePT & ATIS \\
\midrule
Theoretical Privacy Bounds & 16 & $\varepsilon$-Differential Privacy: formal upper bound on leakage; smaller $\varepsilon$ implies stronger protection. 

\(\Pr[M(D)] \le e^{\varepsilon}\Pr[M(D')]\) & \citet{chen_customized_2023} & CusText & SST-2 \\
\bottomrule
\end{tabular}
\caption{Example privacy evaluation metrics by objective. Each row summarizes one representative paper, including the metric it reports, its anonymization method, dataset, and a brief note. \# is the number of surveyed papers reporting metrics in each objective.}
\label{tab:rep_metric_by_objective}
\end{table*}

\subsubsection{Identifier Removal Effectiveness}
\label{sec:identifier}

Identifier removal effectiveness asks whether anonymized text still exposes directly identifying information, such as names, addresses, or contact details that should have been masked during anonymization. The most common evaluation approach compares detected spans against gold standard annotations using token-level or span-level precision, recall, and $F_{1}$ scores \citep{hassan_automatic_2019, manzanares-salor_automatic_2022, manzanares-salor_evaluating_2024, papadopoulou_bootstrapping_2022, papadopoulou_neural_2022, ribeiro_incognitus_2023, iwendi_n-sanitization_2020, dou_reducing_2024, kleinberg_textwash_2022, arranz_mapa_2022}. These metrics dominate because they clearly reflect two types of failure: masking too little (false negatives) and masking too much (false positives) \citep{lison_anonymisation_2021}.

Metrics vary in how they evaluate disclosure. For example, entity-level recall treats an identifier as successfully protected only if all its mentions are masked across a document or corpus, while other metrics distinguish between identifiers that uniquely point to individuals (e.g., names) and those that may only do so in combination (e.g., age or ZIP code) \citep{pilan_text_2022}.
To handle cases where the predicted masked span does not exactly match the annotated boundary—either by masking too little or too much—tagging schemes (IOB-Exact or IOB-Partial) have been used to address partial masking and boundary mismatch \citep{lison_anonymisation_2021}.

Approximate-match metrics credit redactions that are incomplete but still effective, such as changing “John Smith” to “Jonathan” or using paraphrases. Edit-distance-based scores (e.g., Levenshtein Recall) and token-level lexical divergence quantify how different the anonymized span is from the original \citep{alves_anonymization_2024, xin_false_2025}. At a higher semantic level, metrics like PRIVACY\_NLI ask whether the anonymized sentence still implies the original using textual entailment models, while SPRIVACY reports human judgments of whether personal information remains \citep{huang_nap2_2024}.

Together, these metrics form a progression from surface-level removal to deeper notions of semantic obfuscation. While span-level $F_{1}$ remains the most common metric, newer work shows that semantic or corpus-level assessments may better capture residual privacy risks, especially in cases when anonymization involves rewriting rather than redaction. Importantly, identifier-only metrics may underestimate leakage: successfully masking names does not guarantee protection if the text still allows an individual to be inferred through other cues, which motivates more robust evaluation frameworks under different notions of privacy.

\subsubsection{Dataset Membership}

Dataset membership metrics assess whether an adversary can determine if a specific record was part of the data used to train or generate an anonymized output. This metric is strongly tied to the concept of privacy in models and is most commonly used to assess synthetic data generation approaches, as most redaction-based approaches would trivially fail this test. When the notion of a ``record'' is well defined—as in the entire clinical record for a patient—successfully hiding membership may offer broad protections \citep{Salem2023}.

Standard evaluations use shadow or reference models to estimate membership inference accuracy, $F_1$, or AUC \citep{arnold_guiding_2023, el_kababji_evaluating_2023}. Variants include confidence-threshold and entropy-threshold attacks on privatized embeddings, with success rate indicating leakage \citep{du_sanitizing_2023}. However, membership inference attacks remain fragile and context-sensitive: their performance is highly influenced by attack design, dataset construction, and the nature of the reference data \citep{naseh_synthetic_2025, duan2024membership}.

While membership inference traditionally focuses on entire data points (e.g., if a full document was included in a dataset used to train a synthetic data generator), work focused on synthetic text generation specifically has also evaluated privacy through canary-injection experiments \citep{carlini_secret_sharer_canary, yue_synthetic_2023, ramesh-etal-2024-evaluating}, which assess memorization and leakage through a partial notion of membership. In canary experiments, unique phrases are inserted into the data used to train synthetic text generators. The leakage rate and perplexity rank of these ``canaries'' serve as indicators of the risk of data leakage, e.g., a low perplexity indicates a canary was likely a member of the training data \citep{carlini_secret_sharer_canary}. 

Dataset membership testing is common in evaluations of generation methods that claim differential privacy guarantees \citep{arnold_guiding_2023, du_sanitizing_2023, yue_synthetic_2023}.  These studies often pair empirical leakage measures with formal $(\varepsilon, \delta)$ budgets to assess whether theoretical protections translate into practical robustness.

\subsubsection{Attribute Inference Risk}
\label{sec:attribute}

Attribute inference metrics evaluate whether sanitized text still reveals sensitive traits, that is, can a reader infer gender, age, or diagnosis more accurately than chance? These metrics are often used to evaluate rewriting-based methods \citep{meisenbacher_dp-mlm_2024, meisenbacher_thinking_2024} and synthetic text generation approaches \citep{wang_differentially_2023, wang_promptehr_2022}.
Most published attacks pursue attributes of the text author: reviewer gender or age in Trustpilot, political leaning in tweets, stylistic cues in blog posts \citep{meisenbacher_dp-mlm_2024, chim_evaluating_2025}.  A smaller but growing line of work targets attributes of people mentioned within the text, for example, patient sex or comorbidities in synthetic EHRs generated by DP-RVAE or PromptEHR \citep{wang_differentially_2023, wang_promptehr_2022}.

A common approach is to train classifiers on both original and anonymized text and compare their ability to predict protected traits. A drop in accuracy or $F_{1}$ is interpreted as evidence of improved privacy. For example, \citet{meisenbacher_dp-mlm_2024} uses Privacy $F_{1}$ under static and adaptive attackers to quantify how retraining DP-MLM rewrites limits attribute leakage. Multi-attribute settings extend this to keyword-inference accuracy, Gender-$F_{1}$, and Age-$F_{1}$ in synthetic EHRs \citep{wang_differentially_2023, meisenbacher_thinking_2024}.

Beyond individual attribute prediction, \citet{el_kababji_evaluating_2023} model sequential attacks in which an adversary first links synthetic clinical trial records to real patients and then predicts sensitive attributes such as tumor grade. These approaches capture different facets of attribute leakage and can be applied to both token and embedding-level representations.

These metrics are valuable for quantifying residual leakage that might persist even after identifiers are removed. However, a drop in inference accuracy does not guarantee that private attributes are fully protected \citep{du_sanitizing_2023, chim_evaluating_2025}. Attribute inference thus plays a complementary role in privacy evaluation: it highlights forms of privacy leakage that are not captured by identifier masking alone, but does not ensure broader protection on its own against reconstruction or membership disclosure.

\subsubsection{Reconstruction Attacks}
\label{sec:recon}

Reconstruction attacks pose a different question: after anonymization, can an adversary re-create verbatim or near-verbatim portions of the original document, and thus link them back either to the author or to the individuals mentioned? Even with names removed, rare phrases or consistent style can suffice for re-identification.

The most widely reported reconstruction metrics operate at the document level. Retrieval-based metrics (e.g., BM25, Jaccard, or ensemble linking) count how often an anonymized text's original counterpart is retrieved from the candidate pool, exposing residual uniqueness in wording or topic \citep{xin_false_2025, ben_cheikh_larbi_clinical_2023, morris_unsupervised_2022}. In clinical domains, manual re-identification studies simulate the process by which humans might trace rewritten notes back to the patients described \citep{casula_dont_2024}.

More automated approaches for re-identification include bounding worst-case leakage rates across tokens \citep{tong_vulnerability_2025}, and estimating how easily masked tokens are guessed by models like BERT \citep{chen_customized_2023}. Other metrics highlight unique or memorized content: span surprisal \citep{papadopoulou_neural_2022}, plausible-deniability set size \citep{yue_differential_2021}, ROUGE overlap \citep{zecevic_generation_2024}, and rare-token counts \citep{meisenbacher_comparative_2024}.

Together, these metrics range from coarse retrieval to fine-grained content recovery. Choosing among them depends on whether the primary concern is full-document retrieval or recovery of sensitive snippets, and whether the at-risk party is the author of the text, the individual described, or both.

\subsubsection{Semantic Inference Risk}
\label{sec:semantic}

While reconstruction metrics focus on verbatim overlap, semantic inference metrics ask a broader question: does the anonymized text still convey the same meaning as the original? If so, an adversary may infer sensitive information, even in the absence of explicit identifiers. Metrics focused on semantic inference differ from those focused on attribute inference in that they primarily assess similarity between original and anonymized text, and do not necessarily target the prediction of specific personal traits. Instead, they flag risks when anonymized content retains enough topical, relational, or narrative structure to support inference.

Most evaluations begin with embedding-based similarity. SBERT cosine scores are commonly used to quantify alignment between original and anonymized text \citep{meisenbacher_1-diffractor_2024}. To move beyond raw cosine scores, \citet{xin_false_2025} introduce two refinements: a lexical divergence score, which filters out superficial rewording, and a semantic alignment score, which uses language model prompts to judge factual consistency. Both metrics help identify cases where surface anonymization fails to hide deeper meaning, especially in clinical contexts where sensitive events remain recognizable.

When dense encoders are unavailable, simpler lexical metrics provide a coarse but practical alternative. \citet{igamberdiev_dp-bart_2023} reinterpret corpus-level BLEU as a privacy indicator: large \(n\)-gram overlap implies that substantial original wording and therefore potential leakage remains. \citet{meisenbacher_comparative_2024} report the Perturbation Percentage (PP), the fraction of tokens altered during anonymization, and show that low PP often aligns with successful author-attribute inference.

These metrics vary in granularity and precision, but all reflect a central tradeoff: preserving utility often means preserving meaning, which may leave privacy at risk. For applications where downstream utility is paramount, practitioners may tolerate relatively high similarity scores, whereas in high-sensitivity domains, such as clinical text, even subtle semantic similarities can pose privacy risks.

\subsubsection{Theoretical Privacy Bounds}

In contrast to the more empirical measurements of privacy described in the preceding sections, differentially private methods provide strict mathematical guarantees that set an upper bound on the maximum permissible privacy leakage. The tightness of this bound depends on the user-specified parameters that define the privacy budget, and it can be reported alongside other empirical measures. As such, we regard these theoretical bounds as distinct metrics that directly quantify the extent of privacy protection.

Differentially private (DP) methods generally involve the addition of noise to model representations or gradient updates to reduce the risk of membership inference. The level of noise added is carefully calibrated to adhere to the specified privacy budget, typically formalized by ($\varepsilon$, $\delta$) differentially private guarantees.
Both text synthesis and text rewriting approaches have incorporated DP guarantees, including SANTEXT, DP-BART, DP-MLM, and DP-RVAE, and papers typically report results under varying levels of $\varepsilon$ \citep{yue_differential_2021, igamberdiev_dp-bart_2023, meisenbacher_dp-mlm_2024, wang_differentially_2023,du_sanitizing_2023}.



In addition to global $\varepsilon$ values, some studies analyze privacy at the token level to better understand the behavior of specific mechanisms. The self-substitution rate \(N_{w}\) measures the probability that a token survives the mechanism unchanged, whereas the support size \(S_{w}\) counts how many distinct outputs the mechanism may emit for that token \citep{meisenbacher_1-diffractor_2024, meisenbacher_comparative_2024, arnold_guiding_2023}. These two metrics together characterize the output entropy of the substitution process: when tokens are frequently altered and drawn from a large set of alternatives, an adversary faces greater uncertainty about the original content.

Importantly, theoretical guarantees do not replace empirical testing. They only hold if methods are correctly implemented and the data satisfy the necessary assumptions. 
Furthermore, while theoretical metrics can precisely describe which setup offers better protection, they typically lack human interpretability, e.g., under differential privacy, $\varepsilon=4$ implies better protection than $\varepsilon=8$, but it is not clear what either metric actually means for leakage risks nor which value should be used.
Recent studies report DP parameters alongside reconstruction or membership attack results, enabling readers to verify whether the empirical results respect the advertised guarantees \citep{meisenbacher_1-diffractor_2024, chen_customized_2023, wang_differentially_2023, meisenbacher_dp-mlm_2024, zecevic_generation_2024, arnold_guiding_2023, du_sanitizing_2023, yue_synthetic_2023, wang_promptehr_2022}.

\section{Are current metrics sufficient to meet legal standards?}
\label{sec:legal}

Modern privacy regulations articulate rigorous requirements for anonymization that are not always reflected in current technical evaluations. In this section, we assess whether commonly used evaluation metrics in text anonymization align with the legal definitions, using the two most influential frameworks as case studies: the U.S. HIPAA Privacy Rule and the EU General Data Protection Regulation (GDPR). Drawing from the survey in \Sref{sec:survey}, we analyze where current practices fall short, and what improvements are necessary for legal defensibility.

\subsection{HIPAA: Emphasis on Identifier Removal and Expert Judgment}

The HIPAA Privacy Rule defines two standards for de-identification of data: (1) Safe Harbor, which mandates removal of 18 enumerated identifiers; and (2) Expert Determination, in which a statistical expert attests that the risk of re-identification is ``very small'' given anticipated use \citep{rights_ocr_guidance_2012}.

Identifier removal metrics (\Sref{sec:identifier}) align well with Safe Harbor. These metrics appeared in 15 of the reviewed papers, and directly measure how effectively models detect and mask identifiable tokens.

However, current evaluation datasets are rarely annotated according to HIPAA standards. Annotation of generic named entity types misses more domain-specific identifiers, especially since HIPAA's list includes quasi-identifiers like geographic information and dates.
Entity-level recall metrics \citep{pilan_text_2022} better quantify HIPAA compliance than span-level metrics by requiring consistent masking across contexts, but few evaluations use them.

The Expert Determination pathway implies the need for holistic risk modeling—evaluations that simulate adversarial re-identification or analyze residual inference risks. While attribute inference, reconstruction attacks, and semantic inference risk have the potential to mimic expert determinations, only a few studies attempt such modeling, and very few studies investigate how attack models compare to real experts. Exceptions include human-in-the-loop evaluations, such as the TILD framework \citep{mozes_no_2021}, which uses ``motivated intruder'' tests to assess whether humans can re-identify entities given background knowledge.

While current evaluation metrics cover some aspects of HIPAA, especially Safe Harbor, they fall short of the broader requirements implied by Expert Determination, which demand more comprehensive and adversary-aware assessments.

\subsection{GDPR: Contextual Risk and Semantic Inference}
GDPR requires that anonymized data be such that individuals are ``not identifiable by any means reasonably likely to be used'' by an adversary \citep{european_regulation_2016}. This contextual standard evaluates identifiability not just by direct identifiers but also by semantic clues, auxiliary data, and task-specific inference.

Reconstruction metrics (\Sref{sec:recon}) simulate adversarial behavior and are among the most legally aligned with GDPR. However, most studies adopt a single fixed attacker and rarely vary the knowledge base or background assumptions, limiting their robustness as legal evidence.

Attribute inference metrics (\Sref{sec:attribute}) also relate directly to GDPR concerns, as they measure the extent to which sensitive traits can be recovered from anonymized text. Yet few evaluations test multiple attributes.

Metrics from the semantic inference risk (\Sref{sec:semantic}) category indirectly assess the residual information in the text. High semantic similarity may indicate exposure of sensitive attributes or events. Yet these proxies do not directly evaluate whether an attacker could infer private information, as required under GDPR.

GDPR compliance requires adversarial thinking and evaluation of contextual identifiability. Most current metrics fall short on this front: identifier removal metrics overlook quasi-identifiers and risks of re-identification; Reconstruction metrics are rarely diversified across attack strategies; and semantic similarity scores do not map cleanly onto real-world inference risks. Broader adoption of human-intruder studies and diverse reconstruction attacks and attribute inference probes are needed to bridge this gap.

\section{User-Centered Privacy and Contextual Integrity}
\label{sec:user}

While technical metrics dominate text anonymization research, they often overlook a central question: to what extent do these metrics reflect what people actually care about in privacy? Human-centered literature on human–computer interaction (HCI) and social computing suggests that users' privacy perceptions depend on more than whether names or attributes are masked. Users' privacy expectations are shaped by the information context, agency, and perceived coherence of privatized text. This section explores key themes from user-centered privacy research, identifying gaps between current evaluation practices and the lived concerns of users.

The theory of Contextual Integrity, introduced by \citet{nissenbaum_privacy_2004}, suggests that privacy is not about secrecy or control in the abstract, but about appropriate flows of information: who sends what to whom, under what conditions, and for what purpose. In practice, whether a particular data sharing is acceptable depends on if it aligns with the norms in the associated context. For example, acceptance of COVID-19 contact tracing applications and vaccination-certification systems depends on whether the information flows are bounded by expectations about recipients, use purpose, and retention time, all of which go beyond simply removal of identifiers or risks of re-identification \citep{feng_contextual_2024, zhang_stop_2022}.
Through a user study with 721 participants, \citet{meisenbacher_investigating_2025} show that users care about data sensitivity, mechanism type, and reason for data collection in the specific context of differentially private text, as suggested by contextual integrity theory more generally.



While NLP systems and evaluation practice have minimally drawn from contextual integrity theory, 
HCI studies have leveraged it by treating privacy as alignment between users' disclosure preferences and the contextual demands, rather than as fixed rules or outputs.
Several systems aim to support users in managing what they share, rather than deciding for them. For instance, Rescriber lets users rewrite or hide sensitive parts of their messages to language models, based on what the user feels is appropriate in the moment \citep{zhou_rescriber_2025}. Other tools like CLEAR and Contextual Privacy Policies adapt the way data is handled depending on factors like location, app behavior, or who the recipient is \citep{chen_clear_2025, pan_new_2024}.



Evaluations of privacy in text could similarly integrate context. Currently, metrics focus narrowly on identifier recall, leakage, or attack success, without assessing whether the anonymized text reflects an information flow that is appropriate for the context, whether users feel in control of disclosures, or whether the outputs align with their privacy expectations. As a result, systems may score well on standard benchmarks yet still fail to earn user trust or meet real-world standards of privacy acceptability.
Context-sensitive metrics could entail, for example, explicitly defining the scenario where each metric is appropriate. Future work could also develop new metrics that take context or user preferences as input variables that influence the type of assessment.

\section{Discussion}
\label{sec:discussion}

While our survey focuses on evaluating privacy in text itself, a related line of research concerns the privacy risks of models trained on text, with recent work focusing on large language models (LLMs).
We briefly highlight how our survey can inform research in this setting as well, and generally suggest that better reconciling text and model privacy can advance both areas.

Model privacy literature typically investigates whether trained models can memorize, reveal, or allow inference about sensitive training data \citep{neel_privacy_2024}. Although the evaluation target differs from text anonymization, the two areas share some similar privacy notions, such as membership inference and reconstruction attacks. Specific metrics for model privacy overlap with metrics used to evaluate privacy in synthetic text, including canary attacks and success rate of membership inference attacks, where evaluation often targets the synthetic text generated, not just the output text. In particular, membership inference attacks (MIAs) have been widely studied in both black-box and white-box settings \citep{carlini_membership_2022, shokri_membership_2017}, with recent work adapting them to few-shot and in-context learning \citep{wen_membership_2024, jimenez-lopez_membership_2025}. 

Beyond leakage of training data, \citet{staab_beyond_2024} demonstrate an additional model privacy risk in LLMs specifically: that they can infer sensitive traits through attribute inference attacks. This risk is quantified using metrics like classifier accuracy and profiling success, which also appear in anonymization work \citep{frikha_incognitext_2025}.
Although a privacy risk, the potential for LLMs to be powerful de-anonymizers also offers an opportunity for empirical evaluation: LLMs may serve as strong adversaries in empirically conducting reconstruction attacks, attribute inference risks, and semantic inference risks.

Model privacy literature includes several standardized benchmarks.
\citet{mireshghallah_can_2024} apply theories of contextual integrity to evaluate privacy in terms of normative expectations, echoing similar calls in user-centered anonymization metrics.
PrivLM-Bench evaluates privacy risks such as PII exposure and attribute inference across standardized tasks \citep{li_privlm-bench_2024}. Probing tools like ProPILE and targeted black-box attacks offer practical approaches to assess leakage without requiring internal model access \citep{kim_propile_2023, abascal_tmi_2024}. These methods highlight how information can leak through paraphrases or semantic proxies, a challenge also present in text anonymization. As privacy risks in NLP span both model behavior and textual output, bridging the two literatures could support more robust and transparent evaluation frameworks. These would incorporate attacker simulations, contextual analysis, and metrics grounded in real-world privacy concerns.

\section{Recommendations and Open Challenges}
\label{sec:rec}

Our survey reveals several gaps in current privacy evaluation practices for text anonymization and highlights opportunities for future work. In this section, we synthesize key takeaways into actionable recommendations and outline open research directions for building more robust and comparable evaluation frameworks.

\paragraph{Align metrics with stated goals.}
Privacy metrics should reflect the intended privacy guarantees of a method. For example, approaches designed to minimize re-identification risk should not be evaluated solely with identifier-removal F1 scores, which ignore indirect leakage. Similarly, methods that aim to reduce semantic inference should adopt task-specific probes or classifier-based evaluations, not just surface similarity metrics. Articulating intended use cases and mapping them to appropriate metrics is essential for meaningful evaluation.


\paragraph{Design comparable and use-case-grounded evaluations.}
The field would benefit from standardized evaluation pipelines that apply uniformly across anonymization strategies.
Currently, text anonymization methods are frequently evaluated under different notions of privacy. For example, while redaction approaches are evaluated for identifier removal, synthetic data generation methods are evaluated using membership inference attacks. The lack of standardization makes it difficult to compare the practical usability of these approaches.
Evaluation protocols should be grounded in realistic scenarios and expected use cases, rather than tailored to probing the specific proposed method. 

\paragraph{Support human-centered and context-aware evaluation.}
Current metrics often overlook privacy risks that arise from context or user expectations. Approaches such as motivated intruder tests, where a human tries to re-identify records using web searches or domain knowledge, contextual acceptability judgments, and scenario-based probing can help capture privacy violations not visible through token-level leakage scores. While these methods are expensive, they offer high-fidelity signals that better reflect real-world privacy concerns.

\paragraph{Bridge technical metrics with legal standards.}
Technical evaluations should be interpretable in light of legal definitions of identifiability and risk, recognizing that strong performance on token-level metrics may not satisfy privacy laws or user expectations. Integrating adversarial simulations, auxiliary knowledge tests, and plausibility-based linkage metrics can help ensure evaluations better reflect regulatory expectations. At the same time, current policies often lag behind emerging threats. Over time, robust and transparent evaluation metrics, especially those grounded in real-world risks, should inform the development of improved legal standards and regulatory benchmarks.

\paragraph{Scale and structure human-in-the-loop evaluation.}
Manual re-identification or attribute inference studies offer valuable insights, but are costly and difficult to reproduce. To make them more reproducible and scalable, future work should develop annotation protocols, intruder test guidelines, and hybrid heuristics that combine automation with targeted human review. Establishing norms for reporting such studies would also support transparency and comparison.


By addressing these issues, future research can move toward a more comprehensive, reliable, and socially grounded framework for evaluating privacy in text anonymization.

\section{Conclusion}
Text anonymization remains an essential yet difficult component of privacy-preserving NLP. Our survey identifies six distinct privacy objectives reflected in existing metrics and highlights gaps between current evaluation practices and the broader legal, social, and practical standards that define meaningful privacy protection.

To move toward more rigorous and relevant evaluation, we call for clearer alignment between stated privacy goals and chosen metrics, greater attention to adversarial and contextual risks, and stronger integration of human-centered perspectives. As privacy risks grow with increasingly powerful generative models, a structure and context-aware evaluation framework will be key to ensuring responsible data sharing and model deployment.

\section*{Limitations}
This survey focuses exclusively on post hoc evaluation metrics for privacy in text anonymization. We do not assess the effectiveness of anonymization methods themselves.
We also do not conduct a thorough review of other privacy paradigms, such as model privacy (except where they relate to our work) or federated learning.   

Our inclusion criteria require papers to explicitly report at least one privacy metric, which may bias our sample toward works that adopt quantifiable evaluation practices.
Finally, while we discuss legal and social notions of privacy, our analysis is necessarily interpretive and does not constitute formal legal guidance.

\section*{Acknowledgments}
The authors would like to thank the reviewers for their helpful feedback. This work was supported in part by the AI2050 Fellowship program by Schmidt Sciences.

\bibliography{custom}

\end{document}